\documentclass[10pt,twocolumn,letterpaper]{article}

\usepackage{cvpr}
\usepackage{times}
\usepackage{epsfig}
\usepackage{graphicx}
\usepackage{amsmath}
\usepackage{amssymb}
\usepackage{multirow}

\usepackage[breaklinks=true,bookmarks=false]{hyperref}

\cvprfinalcopy 

\def\cvprPaperID{****} 

\begin{document}

\title{PointTrack++ for Effective Online Multi-Object Tracking and Segmentation}

\author{Zhenbo Xu\textsuperscript{\rm 1},
Wei Zhang\textsuperscript{\rm 2},
Xiao Tan\textsuperscript{\rm 2},
Wei Yang\textsuperscript{\rm 1}\thanks{Corresponding author. Email: qubit@ustc.edu.cn},
Xiangbo Su\textsuperscript{\rm 2},
Yuchen Yuan\textsuperscript{\rm 2},\\
Hongwu Zhang\textsuperscript{\rm 2},
Shilei Wen\textsuperscript{\rm 2},
Errui Ding\textsuperscript{\rm 2},
Liusheng Huang\textsuperscript{\rm 1}
\\
\textsuperscript{\rm 1}University of Science and Technology of China\\
\textsuperscript{\rm 2}Department of Computer Vision Technology (VIS), Baidu Inc., China\\
}

\maketitle

\begin{abstract}
   Multiple-object tracking and segmentation (MOTS) is a novel computer vision task that aims to jointly perform multiple object tracking (MOT) and instance segmentation. In this work, we present PointTrack++, an effective online framework for MOTS, which remarkably extends our recently proposed PointTrack framework. To begin with, PointTrack adopts an efficient one-stage framework for instance segmentation, and learns instance embeddings by converting compact image representations to un-ordered 2D point cloud. Compared with PointTrack, our proposed PointTrack++ offers three major improvements. Firstly, in the instance segmentation stage, we adopt a semantic segmentation decoder trained with focal loss to improve the instance selection quality. Secondly, to further boost the segmentation performance, we propose a data augmentation strategy by copy-and-paste instances into training images. Finally, we introduce a better training strategy in the instance association stage to improve the distinguishability of learned instance embeddings. The resulting framework achieves the state-of-the-art performance on the 5th BMTT MOTChallenge.
\end{abstract}

\section{Introduction}
Multi-object tracking (MOT) is an essential task in computer vision with broad applications such as robotics and video surveillance. It is widely noticed that object detection and association become challenging in crowded scenes where bounding boxes (bboxes) of different objects might overlap heavily. Recently, multi-object tracking and segmentation (MOTS) \cite{voigtlaender2019mots} extends MOT by jointly considering instance segmentation and tracking. 
In addition to bbox annotations, MOTS provides pixel-wise segmentation labels. As segments precisely delineate the visible object boundaries and separate adjacent objects naturally, MOTS not only enables pixel-level analysis but more importantly encourages to learn more discriminative embeddings for instance association based on segments rather than bboxes.

Nevertheless, learning instance embeddings from segments have rarely been explored by current MOTS methods. TRCNN \cite{voigtlaender2019mots} extends Mask-RCNN to jointly process consecutive frames using 3D convolutions and adopts ROI Align to extract instance embeddings in bbox proposals. To focus on the segment area, Porzi \textit{et al.} \cite{porzi2019learning} introduce mask pooling rather than ROI Align for instance feature extraction. However, vanilla 2D or 3D convolutions are harmful for learning discriminative instance embeddings due to inherent large receptive fields. Deep convolutional features not only mix up the foreground area and the background area but also mix up the foreground area of the interested instance and its adjacent instances. Therefore, though current MOTS methods adopt advanced segmentation backbones to extract image features, they fail to learn discriminative instance embeddings which are essential for robust instance association, resulting in limited performances.

In our previous work, we propose a simple yet highly effective method named PointTrack \cite{xu2020Segment} to learn instance embeddings on segments. As bbox-proposal based instance segmentation methods always miss bboxes when instances multiple bboxes are heavily overlapped, PointTrack adopts a proposal-free instance segmentation network \cite{Neven_2019_CVPR} following the encoder-decoder architecture for efficient instance segmentation. Afterward, for each instance, PointTrack regards raw 2D image pixels as un-ordered 2D point clouds and learns instance embeddings on segments in a point cloud processing manner. As the instance embeddings are learned from raw pixels based on predicted segments, the instance segmentation stage and the instance embedding extraction stage are completely decoupled. In this way, different from previous works \cite{voigtlaender2019mots} which requires consecutive frames as inputs, PointTrack enables a more flexible training strategy since both image and video level segmentation labels can be used. Built on PointTrack, in this paper, we propose PointTrack++ which improves PointTrack by three modifications. Firstly, based on the observation that the poor seed map prediction results in many false-positives and false-negatives, we replace the seed map branch with the semantic segmentation branch and regard the semantic class confidence as the seed score for pixel selection in inference. Secondly, to create more crowded scenes for instance segmentation training, we introduce the Copy-and-Paste strategy by copying instances with similar lightness to cover instances in training images. Lastly, based on the intuition that larger intra-track-id discrepancy which is beneficial for learning the foreground embeddings is harmful for learning the environment embeddings and the position embeddings, we propose a multi-stage training for learning more discriminative instance embeddings. The resulting framework PointTrack++ ranks first on the official KITTI MOTS leader-board and is the winning solution for 5th BMTT MOTChallenge.

\vspace{-2mm}
\section{Methodology}
In this section, we introduce the framework of PointTrack and three improvements that we made in PointTrack++.

\begin{figure}[!t]
\centering
\includegraphics[width=1.0\linewidth]{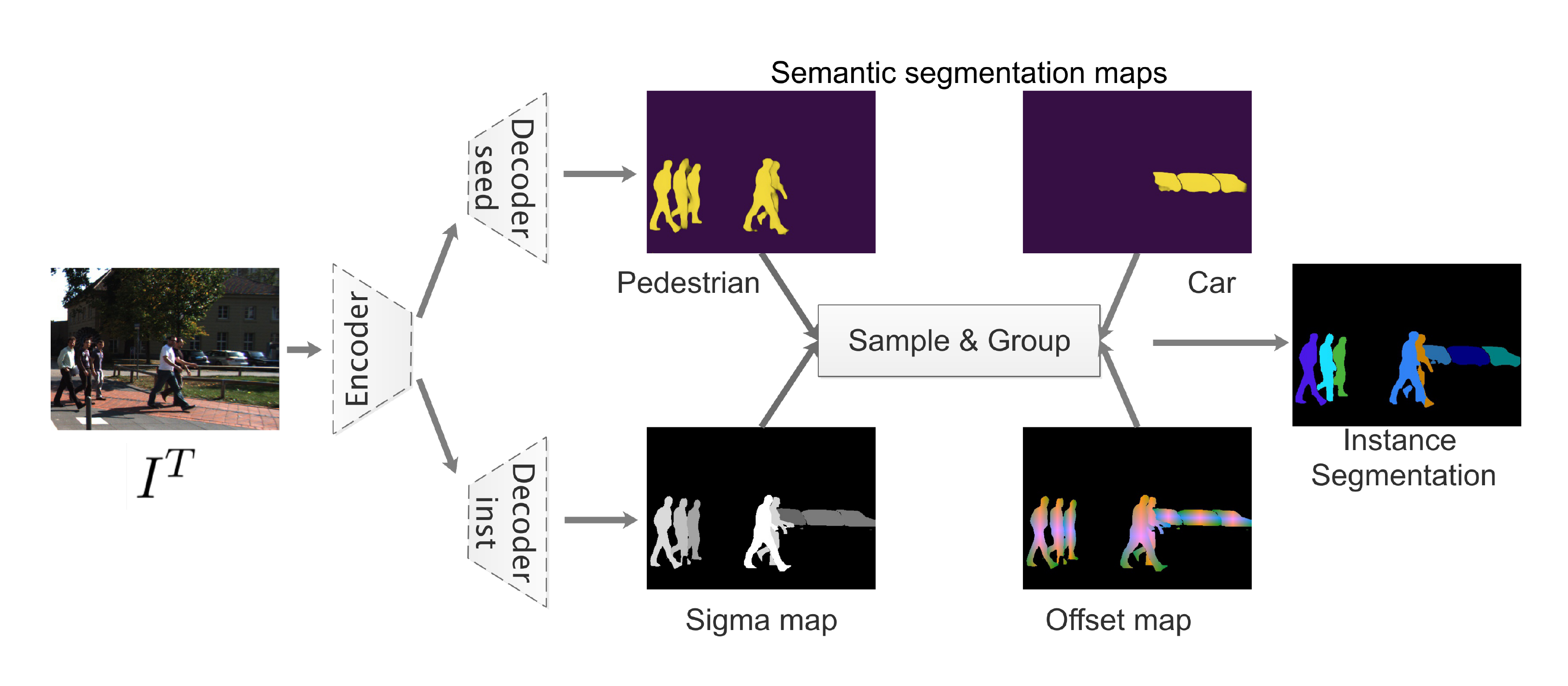}
\caption{Segmentation network of PointTrack++.}
\label{instance_seg}
\vspace{-4mm}
\end{figure}

\begin{figure*}[!t]
\centering
\includegraphics[width=0.6\textwidth]{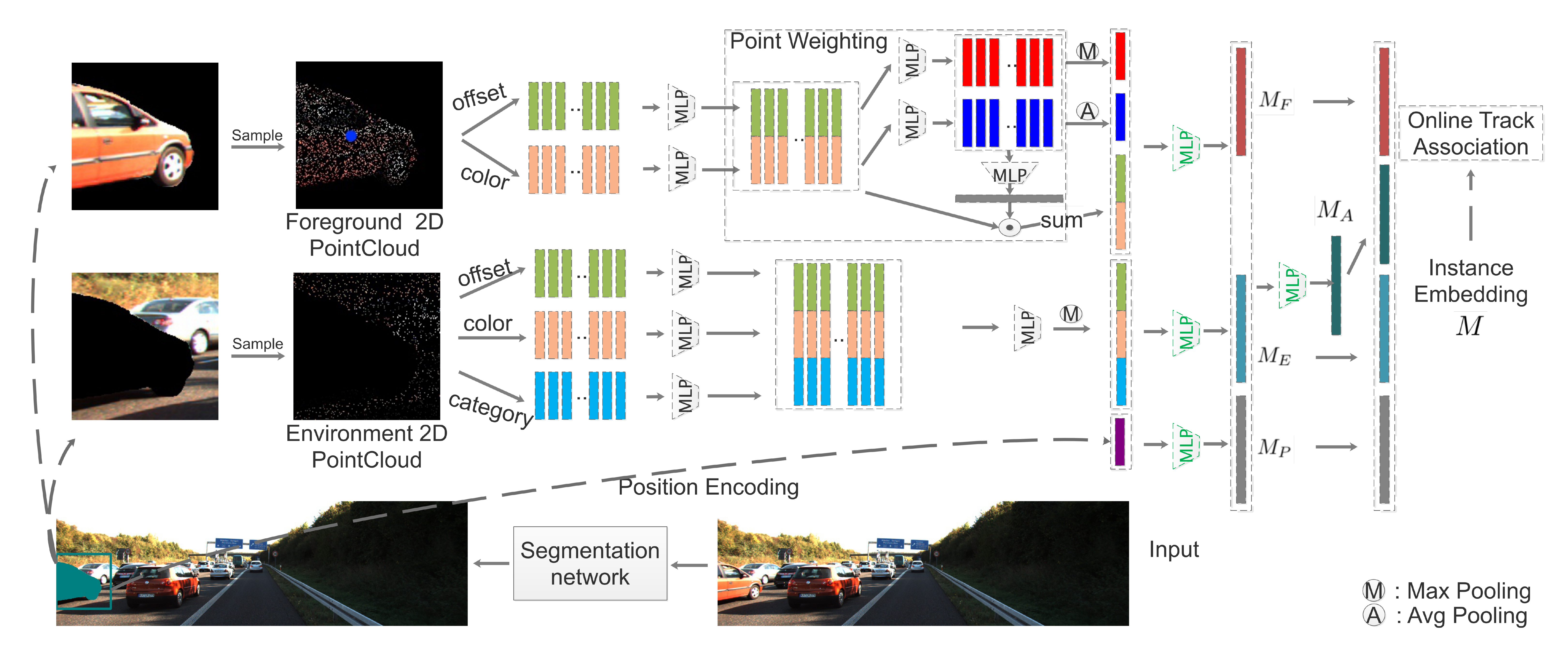}
\caption{Embedding network of PointTrack++. MLP stands for multi-layer perceptron with Leaky ReLU.}
\label{instance_emb}
\vspace{-4mm}
\end{figure*}

\subsection{Overview}

PointTrack \cite{xu2020Segment} contains two major stages including the segmentation stage and the embedding stage. The segmentation network processes the input image and produces the instance segmentation result in a bbox proposal-free manner. Based on the segmentation result, a PointNet-like embedding network is proposed to extract discriminative embeddings for each instance mask.

In the segmentation stage, based on SpatialEmbedding \cite{Neven_2019_CVPR}, PointTrack follows an encoder-decoder structure with two decoders. As shown in Fig. \ref{instance_seg}, given an input image $I^{T}$ at time T, the seed decoder predicts seed maps $S^{T}$ for all semantic classes. Moreover, the inst decoder predicts a sigma map denoting the pixel-wise cluster margin and an offset map representing the pixel-wise normalized vector pointing to its corresponding instance center. Based on the learned clustering margin and normalized vectors, the offsets from the pixel positions in the image plane to its instance center can be computed. In the inference stage, for each semantic class, we recursively group instances by first selecting the pixel with the highest seed value and then grouping nearby pixels to the same instance according to their predicted offsets.

In the embedding stage, following PointTrack \cite{xu2020Segment}, PointTrack++ combines three different data modalities for each sampled pixel and learns context-aware instance embeddings on segments. These three modalities are: (i) Location; (ii) Color; (iii) Category. As shown in Fig. \ref{instance_emb}, for each instance $C$ with its segment $C_s$ and enlarged bbox $\hat{C_b}$, we regard the foreground segment and its environment area as two different 2D point clouds. Afterward, for each point cloud, we uniformly sample points, or say pixels, for feature extraction. Moreover, we also encode the position of $\hat{C_b}$ into the position embedding $M_P$. $M_F$, $M_E$, and $M_P$ denotes the learned foreground embeddings, the environment embeddings, and the position embeddings respectively. Lastly, three embeddings are concatenated and the last MLP is applied to predict the final instance embeddings $M$. Please refer to PointTrack \cite{xu2020Segment} for more details. 

\subsection{Semantic Segmentation Map as Seed Map}
Following SpatialEmbedding \cite{Neven_2019_CVPR}, the original PointTrack \cite{xu2020Segment} creates the Gaussian heat-map as the ground truth of the seed map based on the predicted instance cluster margin. Afterward, the seed map is optimized by the mean squared error of all pixels between the predicted seed map and the Gaussian heat-map. Though the seed loss for foreground pixels has larger weights than background pixels (10 Vs. 1 by default), the predicted seed map, which results in many false-positives and false-negatives in evaluation, is relatively poor. As the seed map is used to sample foreground pixels, we propose to optimize the semantic segmentation map rather than the original seed map. Therefore, we change the seed decoder to the semantic segmentation decoder and introduce Focal loss \cite{lin2017focal} to address the pixel-wise class imbalance.

\subsection{Copy-and-Paste for Data Augmentation}
Unlike cars, the heavily occluded cases are relatively rare in the training set for pedestrians. Moreover, unlike rigid cars, differentiating overlapped non-rigid pedestrians is more challenging. Therefore, we propose the Copy-and-Paste strategy to improve segmentation quality, especially for pedestrians. Fortunately, the precise pixel-wise instance annotations provided by MOTS make Copy-and-Paste convenient and effective. Firstly, we construct a pedestrian database by extracting the pixels and segments of all pedestrians. Then, for instances in each training image, we randomly put pedestrians with a similar lightness from the database on a reasonable position. The resulting realistic training images have more crowded scenes and help PointTrack++ achieve higher segmentation quality.

\subsection{Multi-stage Training for Instance Embedding}
Following PointTrack \cite{xu2020Segment}, the embedding network of PointTrack++ is trained end-to-end on batches of different track ids. Each batch consists of $D$ track ids, each with three crops. In PointTrack, these three crops are selected from three equally spaced frames rather than three consecutive frames to increase the intra-track-id discrepancy and the space is randomly chosen between $1$ and $S$ (set to $10$ by default). 

However, given a large interval between sampled frames, both the environment area and the position of the same instance might change so dramatically that it becomes difficult to differentiate different track ids. Empirically, when the embedding network only learns on the environment 2D point cloud, setting $S$ to a value larger than $2$ makes the embedding network not converge. However, when the embedding network only learns on the foreground 2D point cloud, setting $S$ to a large value such as $12$ helps to achieve higher tracking performance. Therefore, we propose to train $M_F, M_E, M_P$ separately on different $S$ by removing the other two embeddings in training.
Afterward, we fix the parameters of three branches except for the last MLP and learn the aggregated instance embedding $M_A$ by appending an additional MLP layer. The final instance embedding is obtained by concatenating $M_F, M_E, M_P, M_A$.

\vspace{-2mm}
\section{Experiments}
We evaluate our method on the challenging KITTI MOTS benchmark. The main results on the validation set are shown in Table. 1, where we compare PointTrack++ with previous state-of-the-art. Further, we show the comparisons on KITTI MOTS testset between PointTrack++ and other state-of-the-arts. Lastly, we perform ablation study to investigate the contribution of the proposed improvements.

\textbf{Experimental Setup.} Following previous works \cite{voigtlaender2019mots,hu2019learning,luiten2020track}, we focus on sMOTSA, MOTSA, and id switches (IDS). All experiments are carried out on a GPU server with Intel i9-9900X and one TITAN RTX. As PointTrack++ can exploit image-level instance segmentation labels for training, we pre-train the segmentation network on the KINS dataset \cite{qi2019amodal}. Afterward, the segmentation network is fine-tuned on KITTI MOTS for 50 epochs at a learning rate of $5 \cdot 10^{-6}$. The modulating factor of Focal loss is set to $2.0$. During the training of the embedding network, we assign $S$ to $8, 2, 1, 5$ for $M_F, M_E, M_P, M$ respectively. For Copy-and-Paste, the probability of pasting a pedestrian is $0.2$ and $0.5$ for cars and pedestrians respectively. Lastly, for PointTrack++, the input image is up-sampled to twice the original size.

We compare recent works on MOTS: TRCNN \cite{voigtlaender2019mots}, MOTSNet \cite{porzi2019learning}, BePix \cite{sharma2018beyond}, and MOTSFusion (online) \cite{luiten2020track}. TRCNN and MOTSNet perform 2D tracking while BePix and MOTSFusion track on 3D. On KITTI MOTS test set, we compare PointTrack++ with more recent results submitted by participants of 5th BMTT MOTChallenge \footnote{Some methods do not have references as they are not published yet.}.

\begin{table*}[]
\centering
\resizebox{0.8\textwidth}{!}{%
\begin{tabular}{|c|l|l|c|c|c|c|c|c|c|}
\hline
\multirow{2}{*}{Type} & \multicolumn{1}{c|}{\multirow{2}{*}{Method}} & \multicolumn{1}{c|}{\multirow{2}{*}{Det. \& Seg.}} & \multirow{2}{*}{Speed} & \multicolumn{3}{c|}{Cars} & \multicolumn{3}{c|}{Pedestrians} \\ \cline{5-10} 
 & \multicolumn{1}{c|}{} & \multicolumn{1}{c|}{} &  & sMOTSA & MOTSA & IDS & sMOTSA & MOTSA & IDS \\ \hline
2D & TRCNN \cite{voigtlaender2019mots} & TRCNN & 0.5 & 76.2 & 87.8 & 93 & 46.8 & 65.1 & 78 \\ \hline
3D & BePix \cite{sharma2018beyond} & RRC+TRCNN & 3.96 & 76.9 & 89.7 & 88 & - & - & - \\ \hline
2D & MOTSNet \cite{porzi2019learning} & MOTSNet & - & 78.1 & 87.2 & - & 54.6 & 69.3 & - \\ \hline
3D & MOTSFusion \cite{luiten2020track} & TRCNN+BS & 0.84 & 82.6 & 90.2 & 51 & 58.9 & 71.9 & 36 \\ \hline
3D & BePix & RRC+BS & 3.96 & 84.9 & 93.8 & 97 & - & - & - \\ \hline
3D & MOTSFusion & RRC+BS & 4.04 & 85.5 & 94.6 & 35 & - & - & - \\ \hline
2D & PointTrack \cite{xu2020Segment} & PointTrack & \textbf{0.045} & 85.5 & 94.9 & 22 & 62.4 & 77.3 & \textbf{19} \\ \hline
2D & PointTrack++ & PointTrack++ & 0.095 & \textbf{86.81} & \textbf{95.95} & \textbf{17} & \textbf{65.51} & \textbf{81.54} & 26 \\ \hline
\end{tabular}%
}
\label{val_comparison}
\caption{\textbf{Results on the KITTI MOTS validation.} Speed is measured in seconds per frame.}
\vspace{-3mm}
\end{table*}

\begin{table*}[]
\centering
\resizebox{0.7\textwidth}{!}{%
\begin{tabular}{|c|c|c|c|c|c|c|c|c|}
\hline
\multirow{2}{*}{Type} & \multirow{2}{*}{Method} & \multirow{2}{*}{Speed} & \multicolumn{3}{c|}{Cars} & \multicolumn{3}{c|}{Pedestrians} \\ \cline{4-9} 
 &  &  & sMOTSA & MOTSA & IDS & sMOTSA & MOTSA & IDS \\ \hline
2D & TRCNN \cite{voigtlaender2019mots} & 0.5 & 67.00 & 79.60 & 692 & 47.30 & 66.10 & 481 \\ \hline
3D & EagerMOT & - & 74.50 & 83.50 & 457 & 58.10 & 72.00 & 270 \\ \hline
3D & MOTSFusion \cite{luiten2020track} & 0.84 & 75.00 & 84.10 & 201 & 58.70 & 72.90 & 279 \\ \hline
- & Lif\_TS & 1.0 & 77.50 & 88.10 & 183 & 55.80 & 67.70 & \textbf{66} \\ \hline
2D & MCFPA \cite{zhang2008global} & 1.0 & 77.00 & 87.70 & 503 & 67.20 & 83.00 & 265 \\ \hline
2D & PointTrack \cite{xu2020Segment} & \textbf{0.045} & 78.50 & 90.90 & 346 & 61.50 & 76.50 & 176 \\ \hline
3D & LITrk & 0.08 & 79.60 & 89.60 & \textbf{114} & 64.90 & 80.90 & 206 \\ \hline
2D & PointTrack++ & 0.095 & \textbf{82.80} & \textbf{92.60} & 270 & \textbf{68.10} & \textbf{83.60} & 250 \\ \hline
\end{tabular}%
}
\label{test_comparison}
\caption{\textbf{Results on the KITTI MOTS test set.} Speed is measured in seconds per frame.}
\vspace{-3mm}
\end{table*}

\textbf{Results on KITTI MOTS validation.} As the input image is up-sampled, PointTrack++ takes twice the inference time of PointTrack. However, obvious sMOTSA increments of 1.31\% and 3.11\% are observed for cars and pedestrians respectively. It is also worth noting that, on the test set (see Table \ref{test_comparison}), PointTrack++ achieves much larger improvements of 4.3\% and 6.6\% for cars and pedestrians. The steady improvements demonstrate the effectiveness of proposed improvements.

\textbf{Results on KITTI MOTS test set.} To further demonstrate the effectiveness of PointTrack++, we report the evaluation results on the official KITTI test set in Table \ref{test_comparison} where our PointTrack++ currently ranks first \footnote{Please check: \textit{http://www.cvlibs.net/datasets/kitti/eval\_mots.php}}.

\textbf{Ablation Study.} In Table \ref{ablation}, we show the impact of four modifications on performance. `2X' denotes up-sampling the input twice the original size. `Sem' denotes adopting the semantic segmentation map as the seed map. `CP' represents Copy-and-Paste and `Sep' represents the multi-stage training for the embedding network. The first row shows the performance of the original PointTrack. As shown in Table \ref{ablation}, applying `2X' and `Sem' brings a small sMOTSA improvement (0.84\%) for cars. However, a large sMOTSA increment of 2.15\% is observed for pedestrians. Moreover, incorporating Copy-and-Paste into training gives 0.52\% sMOTSA gains for pedestrians. Also, by separately training the embedding network, PointTrack++ achieves 0.86\% higher MOTSA.

\begin{table}[]
\centering
\resizebox{\linewidth}{!}{%
\begin{tabular}{|c|c|c|c|c|c|c|c|c|c|}
\hline
\multicolumn{4}{|l|}{} & \multicolumn{3}{c|}{Cars} & \multicolumn{3}{c|}{Pedestrians} \\ \hline
2X & Sem & CP & Sep & sMOTSA & MOTSA & IDS & sMOTSA & MOTSA & IDS \\ \hline
 &  &  &  & 85.5 & 94.9 & 22 & 62.4 & 77.3 & 19 \\ \hline
v &  &  &  & 86.12 & 94.87 & 19 & 63.78 & 78.28 & 23 \\ \hline
v & v &  &  & 86.34 & 95.14 & 19 & 64.65 & 79.27 & 22 \\ \hline
v & v & v &  & 86.37 & 95.09 & 16 & 65.17 & 81.21 & 23 \\ \hline
v & v & v & v & 86.81 & 95.95 & 17 & 65.51 & 81.54 & 26 \\ \hline
\end{tabular}%
}
\label{ablation}
\caption{\textbf{Ablation study on the impact of modifications.}}
\vspace{-5mm}
\end{table}

\vspace{-2mm}
\section{Conclusions}
In this paper, we present an effective online MOTS framework named PointTrack++. PointTrack++ remarkably extends PointTrack with three major modifications. Through these modifications, PointTrack++ achieves higher segmentation quality and better tracking performance, especially for pedestrians. Extensive evaluations on KITTI MOTS demonstrate the effectiveness of PointTrack++.

\vspace{-2mm}
\section{Acknowledgment}
This work was supported by the Anhui Initiative in Quantum Information Technologies (No. AHY150300).

{\small
\bibliographystyle{ieee_fullname}
\bibliography{egbib}

\begin{thebibliography}{10}\itemsep=-1pt

\bibitem{hu2019learning}
Anthony Hu, Alex Kendall, and Roberto Cipolla.
\newblock Learning a spatio-temporal embedding for video instance segmentation.
\newblock {\em arXiv preprint arXiv:1912.08969}, 2019.

\bibitem{lin2017focal}
Tsung-Yi Lin, Priya Goyal, Ross Girshick, Kaiming He, and Piotr Doll{\'a}r.
\newblock Focal loss for dense object detection.
\newblock In {\em Proceedings of the IEEE international conference on computer
  vision}, pages 2980--2988, 2017.

\bibitem{luiten2020track}
Jonathon Luiten, Tobias Fischer, and Bastian Leibe.
\newblock Track to reconstruct and reconstruct to track.
\newblock {\em IEEE Robotics and Automation Letters}, 2020.

\bibitem{Neven_2019_CVPR}
Davy Neven, Bert~De Brabandere, Marc Proesmans, and Luc~Van Gool.
\newblock Instance segmentation by jointly optimizing spatial embeddings and
  clustering bandwidth.
\newblock In {\em The IEEE Conference on Computer Vision and Pattern
  Recognition (CVPR)}, June 2019.

\bibitem{porzi2019learning}
Lorenzo Porzi, Markus Hofinger, Idoia Ruiz, Joan Serrat, Samuel~Rota Bul{\`o},
  and Peter Kontschieder.
\newblock Learning multi-object tracking and segmentation from automatic
  annotations.
\newblock {\em arXiv preprint arXiv:1912.02096}, 2019.

\bibitem{qi2019amodal}
Lu Qi, Li Jiang, Shu Liu, Xiaoyong Shen, and Jiaya Jia.
\newblock Amodal instance segmentation with kins dataset.
\newblock In {\em Proceedings of the IEEE Conference on Computer Vision and
  Pattern Recognition}, pages 3014--3023, 2019.

\bibitem{sharma2018beyond}
Sarthak Sharma, Junaid~Ahmed Ansari, J~Krishna Murthy, and K~Madhava Krishna.
\newblock Beyond pixels: Leveraging geometry and shape cues for online
  multi-object tracking.
\newblock In {\em 2018 IEEE International Conference on Robotics and Automation
  (ICRA)}, pages 3508--3515. IEEE, 2018.

\bibitem{voigtlaender2019mots}
Paul Voigtlaender, Michael Krause, Aljosa Osep, Jonathon Luiten, Berin
  Balachandar~Gnana Sekar, Andreas Geiger, and Bastian Leibe.
\newblock Mots: Multi-object tracking and segmentation.
\newblock In {\em Proceedings of the IEEE Conference on Computer Vision and
  Pattern Recognition}, pages 7942--7951, 2019.

\bibitem{xu2020Segment}
Zhenbo Xu, Wei Zhang, Xiao Tan, Wei Yang, Huan Huang, Shilei Wen, Errui Ding,
  and Liusheng Huang.
\newblock Segment as points for efficient online multi-object tracking and
  segmentation.
\newblock In {\em Proceedings of the European Conference on Computer Vision
  (ECCV)}, 2020.

\bibitem{zhang2008global}
Li Zhang, Yuan Li, and Ramakant Nevatia.
\newblock Global data association for multi-object tracking using network
  flows.
\newblock In {\em 2008 IEEE Conference on Computer Vision and Pattern
  Recognition}, pages 1--8. IEEE, 2008.

\end{thebibliography}
}

\end{document}